\documentclass{article}
\usepackage{spconf,amsmath,graphicx}
\usepackage{cite}
\usepackage{amsmath,amssymb,amsfonts}
\usepackage{multirow}
\usepackage{algorithm}
\usepackage{algcompatible}
\usepackage{algpseudocode}
\usepackage{textcomp}
\usepackage{xcolor}

\title{Spectral Domain Convolutional Neural Network}
\name{Bochen Guan$^{1,2,\ast}$, Jinnian Zhang$^{2,\ast}$\thanks{$^{\ast}$Authors contributed equally to this work}, William A. Sethares$^{2}$, Richard Kijowski$^{3}$, Fang Liu$^{4}$}
\address{$^{1}$OPPO US Research Center
$^{2}$University of Wisconsin, Madison \\
$^{3}$ New York University
$^{4}$ Harvard University
}
\begin{document}
%
\maketitle
\begin{abstract}
The memory consumption of most Convolutional Neural Network (CNN) architectures grows rapidly with increasing depth of the network, which is a major constraint for efficient network training on modern GPUs with limited memory, embedded systems, and mobile devices. Several studies show that the feature maps (as generated after the convolutional layers) are the main bottleneck in this memory problem. Often, these feature maps mimic natural photographs in the sense that their energy is concentrated in the spectral domain. Although embedding CNN architectures in the spectral domain is widely exploited to accelerate the training process, we demonstrate that it is also possible to use the spectral domain to reduce the memory footprint, a method we call Spectral Domain Convolutional Neural Network (SpecNet) that performs both the convolution and the activation operations in the spectral domain. The performance of SpecNet is evaluated on three competitive object recognition benchmark tasks (CIFAR-10, SVHN, and ImageNet), and compared with several state-of-the-art implementations. Overall, SpecNet is able to reduce memory consumption by about $60\%$ without significant loss of performance for all tested networks. 
\end{abstract}
\section{Introduction}

Training large-scale CNNs becomes computationally and memory intensive 
\cite{pmlr-v97-tan19a, Survey2017,guan2020video}, especially when limited resources are available. Therefore, it is essential to reduce the memory requirements to allow better network training and deployment, such as applying deep CNNs to embedded systems and cell phones. Several studies \cite{Gist} show that the intermediate layer outputs (feature maps) are the primary contributors to this memory bottleneck. Existing methods such as model compression \cite{Quantized_CNN, NIPS2014_5544}
and scheduling 
\cite{pleiss2017memory}, do not directly address the storage of feature maps. By transforming the convolutions into the spectral domain, we target the memory requirements of feature maps.

In contrast to \cite{Gist}, which proposes an efficient encoded representation of feature maps in the spatial domain, we exploit the property that the energy of feature maps is concentrated in the spectral domain. Values that are less than a configurable threshold are forced to zero, so that the feature maps can be stored sparsely. We call this approach the Spectral Domain Convolutional Neural Network (SpecNet). This new architecture incorporates a memory efficient convolutional block in which the convolutional and activation layers are implemented in the spectral domain. The outputs of convolutional layers are equal to the multiplication of non-zero entries of the inputs and kernels. The activation function is designed to preserve the sparsity and symmetry properties of the feature maps in the spectral domain, and also allow effective computation of the derivative for back propagation. 

\section{Related Work}
\textbf{Model compression:} Models can be compressed in several ways including quantization, pruning and weight decomposition.
With quantization, both weights and the feature maps are converted to a limited number of levels. This can decrease the computational complexity and reduce memory cost \cite{Quantized_CNN}. 
Recently, based on the empirical distribution of the weights and feature maps, non-uniform quantization is applied in \cite{Zhang_2018_ECCV} and \cite{sun2016intralayer} incorporates a learnable quantizer for better performance.

Other approaches to model compression include pruning to remove unimportant connections \cite{sturcture_sparsity}
and 
weight decomposition based on a low-rank decomposition (for example, SVD \cite{NIPS2014_5544}) of the weights in the network for saving storage. 


\textbf{Memory Scheduling:} Since the `life-time' of feature maps (the amount of time data 
must be stored) is different in each layer, it is possible to design data reuse methods to reduce memory consumption \cite{pleiss2017memory}. 
A recent approach proposed by \cite{meng2017training} transfers the feature maps between CPU and GPU, which can allow large-scale training in limited GPU memory with a slightly sacrifice in training speed.
\textbf{Memory Efficient CNN Architectures:} By modifying some structures in popular CNN architectures, time or memory efficiency can be achieved. \cite{rota2018place} combines the batch normalization and activation layers together to use a single memory buffer for storing the results. In \cite{Chao_2019_ICCV}, both memory efficiency and low inference latency can be achieved by introducing a constraint on the number of input/output channels. 


\textbf{CNN in the Spectral Domain:} Some pilot studies have attempted to combine Fast Fourier Transform (FFT) or Wavelet transforms with CNNs \cite{FCNN, Wavelet_CNN}. However, most of these works aim to make the training process faster by replacing the traditional convolution operation with the FFT and the dot product of the inputs and kernel in spectral domain. 
These methods do not attempt to reduce memory, and several approaches (such as the Wavelet CNN) require more memory in order to achieve competitive performance. 

\section{SpecNet}
\label{SpecNet_strcuture}

The key idea of SpecNet rests on the observation that feature maps, like most natural images, tend to have compact energy in the spectral domain. The compression can be achieved by retaining non-trivial values while zeroing out small entries. A threshold ($\beta$) can then be applied to configure the compression rate where larger $\beta$ values result in more zeros in the spectral domain feature maps.
 Therefore, SpecNet represents a new design of the network architecture on convolution, tensor compression, and activation function in the spectral domain and can achieve memory efficiency in both network training and inference.
 
\textbf{Compression in Feature Maps:} Consider 2D-convolution with a stride of 1. In a standard convolutional layer, the output is computed by
\begin{equation}
    y(i,j)=x*k=\sum_{m=0}^{M-1}\sum_{n=0}^{N-1}x(m,n) \mbox{ } k(i-m, j-n),
\end{equation}
where $x$ is an input matrix of size $(M,N)$; $k$ is the kernel with dimension $(N_k,N_k)$. 

The output $y$ has dimensions $(M', N')$, where $M'=M+N_k-1$ and $N'=N+N_k-1$. This process involves $\mathcal{O}(M'N'N_k^2)$ multiplications. Its corresponding spectral form is
\begin{equation}
    Y=X\odot K \label{Y},
\end{equation}
where $X$ is the transformed input in the spectral domain by FFT $X=\mathcal{F}(x)$, and $K$ is the kernel in the spectral domain, $K=\mathcal{F}(k)$.  $\odot$ represents element-wise multiplication, which requires equal dimensions for $X$ and $K$. Therefore, $x$ and $k$ are zero-padded to match their dimensions $(M', N')$. Since there are various hardware optimizations for the FFT \cite{FCNN}, it requires $\mathcal{O}(M'N'\log(M'N'))$ complex multiplications. The computational complexity of (\ref{Y}) is $\mathcal{O}(M'N')$ and so the overall complexity in the spectral domain is $\mathcal{O}(M'N'\log(M'N'))$. Depending on the size of the inputs and kernels, SpecNet can also have a computational advantage over spatial convolution in some cases \cite{FCNN}.

The compression of $Y$ involves a configurable threshold $\beta$, which forces entries in $Y$ with small absolute values (those less than $\beta$) to zero. This allows the thresholded feature map $\hat{Y}$ to be sparse and hence to store only the non-zero entries in $\hat{Y}$, thus saving memory. 


The backward propagation for CNN in the spectral domain is studied in \cite{ayat2019spectral}. Since the feature maps are stored sparsely in the forward propagation, the gradients calculated in the backward propagation are approximations to the true gradients. Therefore, although increasing $\beta$ can save more memory, the introduced approximation error will affect the convergence rate or accuracy.
After the gradient update of kernel matrix in the spectral domain, 
it is further converted
to the spatial domain using the IFFT to save kernel storage. 

A more general case of 2D-convolution with arbitrary integer stride can be viewed as a combination of 2D-convolution with stride of 1 and uniform down-sampling, which can also be implemented in the spectral domain \cite{FCNN}.

\textbf{Activation Function for Symmetry Preservation:} 
In SpecNet, the activation function for the feature maps is designed to perform directly in the spectral domain. For each complex entry in the spectral feature map, 
\begin{equation}
    f(a+ib)= h(a)+ig(b) \label{f}
\end{equation}
where 
\begin{equation}
    h(x)=g(x)=\tanh(x)=\frac{e^x-e^{-x}}{e^x+e^{-x}}.
    \label{tanh}
\end{equation}
The $\tanh$ function is used in (\ref{f}) as a proof-of-concept design for this study. Other activation functions may also be used, but must fulfill the following: 1) They allow inexpensive gradient calculation. 2) Both $g(x)$ and $h(x)$ are monotonic nondecreasing. 3) The functions are odd, i.e. $g(-x)=-g(x)$. 

The first and second rules are standard requirements for nearly all popular activation functions used in modern CNN design. The third rule in SpecNet is applied to preserve the conjugate symmetry structure of the spectral feature maps so that they can be converted back into real spatial features without generating pseudo phases. The 2D FFT is
\begin{equation} \nonumber
\!\!\! X(p,q)=\mathcal{F}(x)=\sum_{m=0}^{M+N_k-2}\sum_{n=0}^{N+N_k-2}w_M^{pm}w_N^{qn}x(m,n)
\end{equation}
where $w_M=e^{-2\pi i/(M+N_k-1)}$, $w_N=e^{-2\pi i/(N+N_k-1)}$. If $x$ is real, i.e., the conjugate of $x$ is itself ($\bar{x}=x$), and
\begin{align}
&X(M+N_k-1-p_0,N+N_k-1-q_0) \nonumber \\
    =&\sum_{m=0}^{M+N_k-2}\sum_{n=0}^{N+N_k-2}w_M^{(M+N_k-1-p_0)m}w_N^{(N+N_k-1-q_0)n}x(m,n)
    \nonumber \\
    =&\sum_{m=0}^{M+N_k-2}\sum_{n=0}^{N+N_k-2}w_M^{-p_0m}w_N^{-q_0n}x(m,n)=\overline{X(p_0, q_0)}\nonumber\
\end{align}

Therefore, $g(x)$ must be odd to retain the symmetry structure of the activation layer to ensure that

\begin{align}
f(\overline{a+ib})&=h(a)+ig(-b)=h(a)-ig(b)=\overline{f(a+ib)} \nonumber
\end{align}
The complete forward propagation of the convolutional block in the spectral domain (including convolution and activation operations) is shown in Algorithm 1.



\begin{algorithm}[tbp]
\textbf{Input:} feature maps $x$ from the last layer with size of $M\times N$; kernel $k$ ($N_k\times N_k$); threshold $\beta$.

\begin{algorithmic}[1]
\If{$x$ in the spectral domain}
\State Set $M'=M$, $N'=N$ and $X=x$.
\Else{
\State Set $M'=M+N_k-1$ and $N'=N+N_k-1$.
\EndIf
\For{$i=1$ to $M'$}
\For{$j=1$ to $N'$}
\If{$X$ is None}
\State $\hat{x}(i,j)=x(i,j)$ if $i\leq M$ and $j\leq N$, and $\hat{x}(i,j)=0$ otherwise.
\EndIf
\State $\hat{k}(i,j)=k(i,j)$ if $i\leq N_k$ and $j\leq N_k$, and $\hat{k}(i,j)=0$ otherwise.
\State Calculate $K=\mathcal{F}(\hat{k})$ and $X=\mathcal{F}(\hat{x})$ if $X$ is None.
\EndFor
\EndFor
}
\State Calculate $Y$ after convolution according to (\ref{Y}).

\State Obtain $\hat{Y}$ where $\hat{Y}(i,j)=Y(i,j)$ if $|Y(i,j)|>\beta$, and $\hat{Y}(i,j)=0$ otherwise.

\State Get $Z=f(\hat{Y})$ where $f$ is defined in (\ref{f})
\end{algorithmic}

\textbf{Output:} The feature map in the spectral domain: $Z$.

\caption{Forward propagation of the convolutional block in SpecNet}
\end{algorithm}





\section{Experiments}
\label{sec:pagestyle}

The feasibility of SpecNet is demonstrated using three benchmark datasets: CIFAR, SVHN and ImageNet, and by comparing the performance of SpecNet implementations of several state-of-the-art networks. All the networks were trained by stochastic gradient descent (SGD) with a batch size of 128 on CIFAR, SVHN and 512 on ImageNet. The initial learning rate was set to 0.02 and was reduced by half every 50 epochs. The momentum of the optimizer was set to 0.95 and a total of 300 epochs were trained to ensure convergence. 



SpecNet is evaluated on four widely used CNN architectures including LeNet \cite{lecun1995comparison}, AlexNet \cite{NIPS2012_4824}, VGG \cite{simonyan2014very} and DenseNet \cite{huang2017densely}. We use the prefix `Spec' to stand for the SpecNet implementation of each network. To ensure fair comparisons, the SpecNet networks used identical network hyper-parameters as the native spatial domain implementations. 


\subsection{Results on CIFAR and SVHN}
\begin{figure}[htbp]
    \centering
    \includegraphics[scale=0.145]{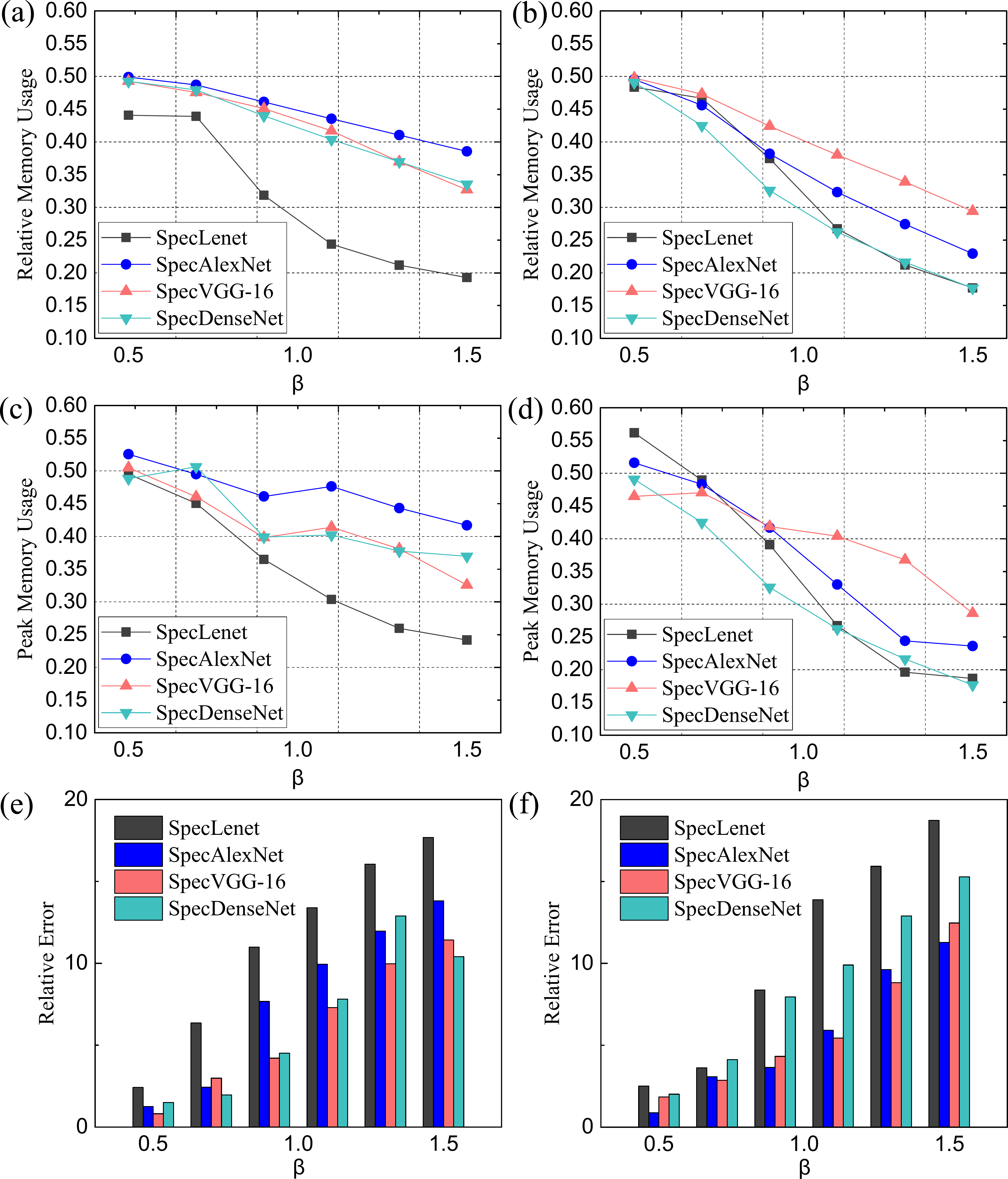}
    \caption{Memory consumption and testing performance of SpecNet versions of AlexNet, VGG, and DenseNet \cite{lecun1995comparison,NIPS2012_4824,simonyan2014very,huang2017densely,wang2017sort} compared to the originals on two datasets. (a)(b)(c) show relative memory consumption and (d)(e)(f) show relative error of SpecNets tested on CIFAR-10 and SVHN.}
    \label{RMS}
\end{figure}

\textbf{Memory}. Fig. \ref{RMS}(a)(b) and (c)(d) compare the average memory usage and peak memory usage of the SpecNet implementations of four different networks over a range of $\beta$ values from 0.5 to 1.5. We quantify the relative memory consumption and accuracy as the memory (accuracy) of SpecNet divided by the memory (accuracy) in the original implementations. When compared with their original models, all SpecNet implementations of the four networks can save at least $50\%$ memory with negligible loss of accuracy, indicating the feasibility of compressing feature maps within the SpecNet framework. With increasing $\beta$ value, all models show reduction in both average accuracy and peak memory usage. The rates of memory reduction are different among the different network architectures, which is likely caused by differences in the feature representations of the various network designs. 

\textbf{Accuracy}. Fig. \ref{RMS}(e)(f) compare the error rates of the SpecNet implementations of the three different networks over a range of $\beta$ values from 0.5 to 1.5. While SpecNet typically compresses the models, there is a penalty in the form of increased error in comparison to the original model with full spatial feature maps. The average accuracy of SpecLeNet, SpecAlexNet, SpecVGG, and SpecDenseNet can be higher than $95\%$ when $\beta$ is smaller than $1.0$.

\begin{figure}[tbp]
    \centering
    \includegraphics[scale=0.31]{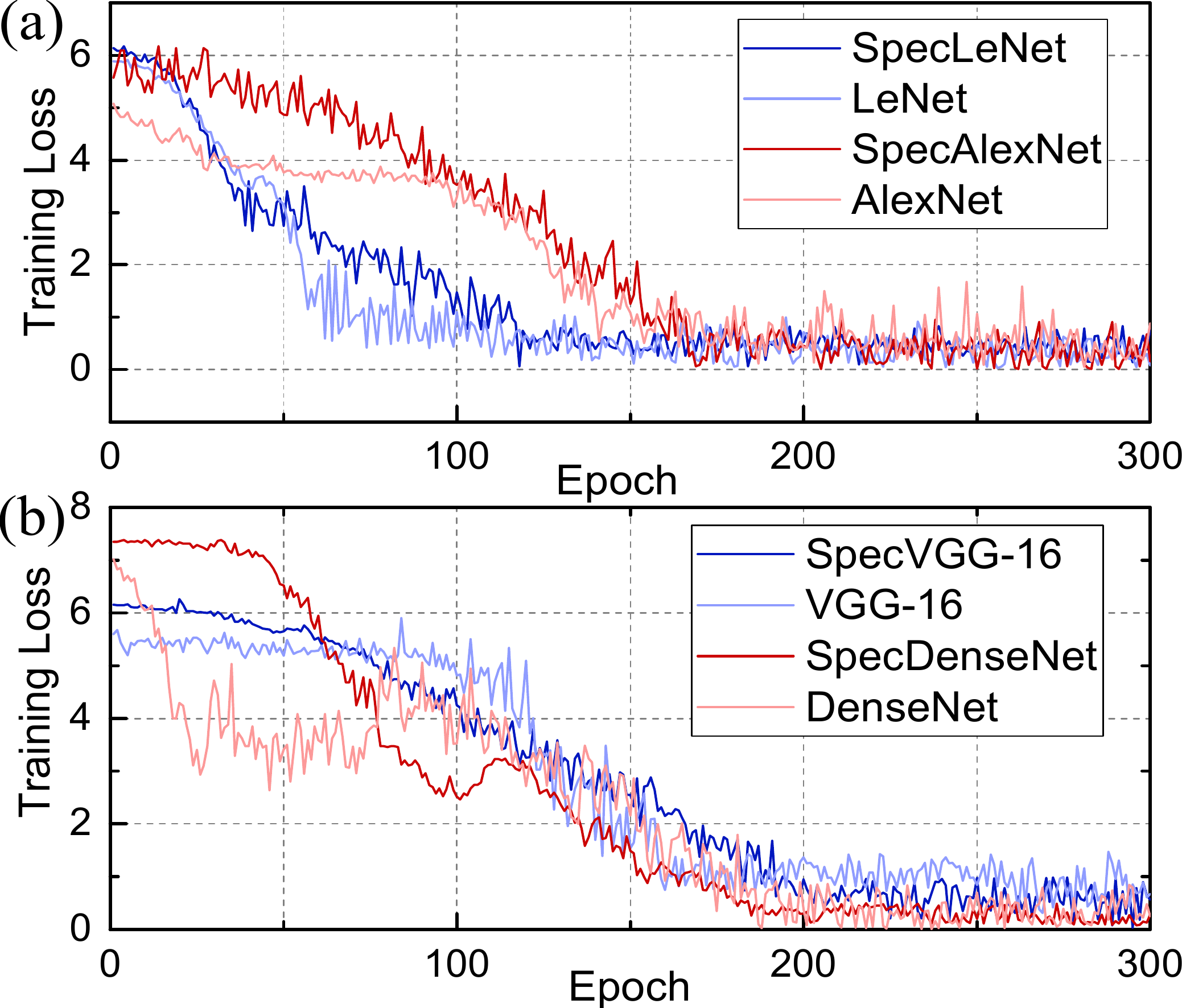}
    \caption{Training curves of SpecNet comparing with , AlexNet, AlexNet, VGG-16 and DenseNet on CIFAR-10 dataset.}
    \label{convergencve}
\end{figure}

Fig.~\ref{convergencve} shows the training curve of the SpecNet implementations of four different networks with their implementations in spatial domain. From the training curve, the SpecNet implementations converge with similar rates as the implementations in the spatial domain. Section \ref{SpecNet_strcuture} also showed that the computation speed is related to the size of the inputs and kernels, and SpecNet is advantageous in some cases. Therefore, training speed of SpecNet is comparable with the network when implemented in the spatial domain, but with the added benefit of memory efficiency. 

Table \ref{tab:Comparison} shows a comparison between SpecNet and other recently published memory-efficient algorithms. The experiments investigate memory usage when training VGG and DenseNet on the CIFAR-10 dataset. SpecNet outperformed all the listed algorithms and resulted in the lowest memory usage while maintaining high testing accuracy. It is notable that SpecNet is independent of the methods listed in the table, and these techniques may be applied in tandem with SpecNet to further reduce memory consumption. 

\begin{table}[tbp]
\centering
\caption{Comparison of relative memory usage for different memory efficient implementations applied to VGG and DenseNet. All the methods are tested on CIFAR-10.}
\scalebox{0.71}{
\begin{tabular}{c|c|c|c|c} 
\hline
\multirow{2}{*}{Model}     & \multicolumn{2}{c|}{VGG-16 (\%)} & \multicolumn{2}{c}{DenseNet (\%)}  \\ 
\cline{2-5}
                           & Memory & Accuray      & Memory  & Accuracy       \\ 
\hline
INPLACE-ABN \cite{rota2018place}               & 52.1         & 91.4        & 58.0         & 92.9          \\
Chen Meng et al. \cite{meng2017training}           & 65.6         & \textbf{92.1}        & 55.3        & 93.2          \\
Efficient-DenseNets \cite{pleiss2017memory} & N/A          & N/A          & 44.3        & \textbf{93.3}          \\
Nonuniform Quantization \cite{sun2016intralayer}   & 80.0          & 91.7        & 77.1        & 92.2          \\
LQ-Net  \cite{Zhang_2018_ECCV}                  & 67.6        & 91.9        & 64.4       & 92.3          \\
HarDNet \cite{Chao_2019_ICCV}              & 46.3        & \textbf{92.1}        & 44.2        & \textbf{93.3}          \\
vDNN \cite{rhu2016vdnn}                      & 54.1   & \textbf{92.1}    & 59.2 & \textbf{93.3}             \\
SpecNet & \textbf{37.0}         & 91.8        & \textbf{37.0}         & 92.5          \\
\hline
\end{tabular}}
\label{tab:Comparison}
\end{table}

\subsection{Results on ImageNet}

We evaluated SpecNet for AlexNet, VGG, and DenseNet on the ImageNet dataset with the $\beta$ value set to $1.0$. We retain the same methods for data preprocessing, hyper-parameter initialization, and optimization settings. Since there is no strictly equivalent batch normalization (BN) method in spectral domain, we remove the BN layers and replace its convolutional layer with the convolutional block in SpecNet (Algorithm 1), keeping all the other experimental settings the same.

\begin{table}[htbp]
\centering
\caption{Memory consumption and testing performance of SpecNet compared with AlexNet, VGG, and DenseNet.}
\scalebox{0.65}{
\begin{tabular}{c|cc|cc} 
\hline
&
\multicolumn{2}{c}{Network Performance}                          & \multicolumn{2}{c}{Memory Comsuption}                                                                                                               \\ 
\hline
                Model
                 &
                 \begin{tabular}[c]{@{}l@{}}
                 Top-1 Val. \\ Acc. (\%)
                 \end{tabular}
                 & 
                 \begin{tabular}[c]{@{}l@{}}
                 Top-5 Val. \\ Acc. (\%) 
                 \end{tabular}
                 & \begin{tabular}[c]{@{}l@{}} Peak Momery \\ Comsuption (\%) \end{tabular} & \begin{tabular}[c]{@{}l@{}} Average Momery \\ Comsuption (\%) \end{tabular}  \\ 
\hline
AlexNet          & 63.3                  & 84.5                 & \multirow{2}{*}{48.1}                                                  & \multirow{2}{*}{49.3}                                                      \\
Spec-AlexNet     & 60.3                 & 84.0                 &                                                                        &                                                                            \\ 
\hline
VGG16            & 71.3                 & 90.7                  & \multirow{2}{*}{42.4}                                                  & \multirow{2}{*}{46.7}                                                      \\
Spec-VGG16       & 69.2                 & 90.5                  &                                                                        &                                                                            \\ 
\hline
DenseNet169      & 76.2                 & 93.2                  & \multirow{2}{*}{36.6}                                                  & \multirow{2}{*}{40.8}                                                      \\
Spec-DenseNet169 & 74.6                 & 93.0                  &                                                                        &                                                                            \\
\hline
\end{tabular}}
\label{ImageNet}
\end{table}

\begin{table}[htbp]
\centering
\caption{Comparison of memory saving for different memory-efficient implementations applied to VGG and DenseNet. All the methods are tested on ImageNet.}
\scalebox{0.55}{
\begin{tabular}{c|ccc|ccc} 
\hline
\multirow{2}{*}{Model}                                              & \multicolumn{3}{c|}{VGG-16}  & \multicolumn{3}{c}{DenseNet}  \\ 
\cline{2-7}
                                                                    & \begin{tabular}[c]{@{}l@{}}
                 Memory \\ ~~ (\%)
                 \end{tabular} 
                   & \begin{tabular}[c]{@{}l@{}}
                 Top-1 Val. \\ Acc. (\%)
                 \end{tabular}
                  & \begin{tabular}[c]{@{}l@{}}
                 Top-5 Val. \\ Acc. (\%)
                 \end{tabular}
                  & \begin{tabular}[c]{@{}l@{}}
                 Memory \\ ~~ (\%)
                 \end{tabular} & \begin{tabular}[c]{@{}l@{}}
                 Top-1 Val. \\ Acc. (\%)
                 \end{tabular}
                 & \begin{tabular}[c]{@{}l@{}}
                 Top-5 Val. \\ Acc. (\%)
                 \end{tabular}   \\ 
\hline
INPLACE-ABN \cite{rota2018place}                 & 64.0       & 70.4  & 89.5 & 57.0       & 75.7 & \textbf{93.2}   \\
Chen Meng et al.\cite{meng2017training}         & 59.4       & 71.0  & 90.1 & 47.0       & 76.0 & 93.0   \\
Efficient-DenseNets \cite{pleiss2017memory}      & N/A         & N/A    & N/A   & 50.7       & 74.5 & 92.5   \\
Nonuniform Quantization \cite{sun2016intralayer} & 76.3       & 69.8 & 87.0 & 75.9       & 73.5 & 91.7   \\
LQ-Net  \cite{Zhang_2018_ECCV}                     & 56.1       & 67.2  & 88.4 & 51.3       & 70.6 & 89.5   \\
HarDNet \cite{Chao_2019_ICCV}                    & 50.1       & \textbf{71.2}  & 90.4 & 47.4       & \textbf{76.4} & 93.1   \\
vDNN \cite{rhu2016vdnn}                           & 57.7       & 71.1  & \textbf{90.7} & 64.9       & 76.0 & \textbf{93.2}   \\
SpecNet                                                             & \textbf{46.7}       & 69.0  & 90.0 & \textbf{40.8}       & 74.6 & 93.0   \\
\hline
\end{tabular}}
\label{tab:Comparison_ImageNet}
\end{table}


We report the validation errors and memory consumption of SpecNet on ImageNet in Table. \ref{ImageNet}. From the experiment, both the average and the peak memory consumption of SpecAlexNet, SpecVGG, and SpecDenseNet are less than half of the original implementations. The maximum memory usage of SpecNet is reduced even more, which is probably due to the extra memory cost of the implementation of convolution by CUDA. 

Table \ref{ImageNet} shows the testing accuracy on ImageNet. Compared with the implementations in the spatial domain, SpecNet has a slight decrease in the top-1 accuracy but almost the same top-5 accuracy ($<96\%$). Thus SpecNet allows a trade-off between accuracy and memory consumption. Users can choose $\beta$ for higher accuracy or for better memory optimization. 

Table \ref{tab:Comparison_ImageNet} shows a comparison between SpecNet and other memory-efficient algorithms on the ImageNet dataset. We investigate memory consumption and accuracy when training VGG and DenseNet. SpecNet shows the largest memory reduction while still maintaining good accuracy. Importantly, our experimental hyper-parameter settings are optimized for the networks in spatial domain. It is likely that more extensive hyper-parameter exploration would further improve
the performance of SpecNet.

\section{Conclusion}
We have introduced a new CNN architecture called SpecNet, which performs both the convolution and activation operations in the spectral domain. We evaluated SpecNet on two competitive object recognition benchmarks, and demonstrated the performance with four state-of-the-art algorithms to show the efficacy and efficiency of the memory reduction. In some cases, SpecNet can reduce memory consumption by $63\%$ without significant loss of performance. 
It is also notable that SpecNet is only focused on the sparse storage of feature maps. In the future, it should be possible to merge other methods, such as model compression and scheduling, with SpecNet to further improve memory usage.

\bibliographystyle{IEEE}
\bibliography{IEEEexample}

\end{document}